\documentclass[twoside,11pt]{article}

\usepackage{graphicx}

\usepackage{jmlr2e}
\usepackage[colorlinks]{hyperref}

\renewcommand{\c}{\mathbf{c}}

\newcommand{\bc}{\mathbf{c}}

\newcommand{\beqn}{\begin{eqnarray}}
\newcommand{\beqnn}{\begin{eqnarray*}}
\newcommand{\eeqn}{\end{eqnarray}}
\newcommand{\eeqnn}{\end{eqnarray*}}

\def \FF {\mathcal{F}}

\def \FF {\mathcal{F}}
\def \no {\Arrowvert}
\def \ind {\hbox{ 1\hskip -3pt I}}
\def \R {\mathbb{R}}

\def \P  {\mathbb{P}} 
\def \R {\mathbb{R}}
\def \E {\mathbb{E}}
\def \N {\mathbb{N}}

\ShortHeadings{Noisy $k$-means}{C. Brunet and S. Loustau}
\firstpageno{1}

\begin{document}

\title{The algorithm of noisy $k$-means}

\author{\name Camille Brunet \email camille.brunet@univ-angers.fr \\
       \addr LAREMA\\
       Universit\'e d'Angers\\
       2 Boulevard Lavoisier,\\
       49045 Angers Cedex, France
       \AND
       \name S\'ebastien Loustau \email loustau@math.univ-angers.fr \\
       \addr LAREMA\\
       Universit\'e d'Angers\\
       2 Boulevard Lavoisier,\\
       49045 Angers Cedex, France}

\editor{}

\maketitle

\begin{abstract}
 In this note, we introduce a new algorithm to deal with finite dimensional clustering with errors in variables. The design of this algorithm is based on recent theoretical advances (see \cite{aoinq,isl}) in statistical learning with errors in variables. As the previous mentioned papers, the algorithm mixes different tools from the inverse problem literature and the machine learning community. Coarsely, it is based on a two-step procedure: (1) a deconvolution step to deal with noisy inputs and (2) Newton's iterations as the popular $k$-means.
\end{abstract}



\section{Introduction}
\label{sec:intro}
One of the most popular issue in data minning or machine learning is to learn clusters from a big cloud of data. This problem is known as clustering or empirical quantization. It has received many attention in the last decades (see \cite{hartigan75} or \cite{graf} for introductory monographs). Moreover, in many real-life situations, direct data are not available and measurement errors may occur. In social science, many data are collected by human pollster, with a possible contamination in the survey process. In medical trials, where chemical or physical measurements are treated, the diagnostic is affected by many nuisance parameters, such as the measuring accuracy of the considered machine, gathering with a possible operator bias due to the human practitionner. Same kinds of phenomenon occur in astronomy or econometrics (see \cite{meister}). However, to the best of our knowledge, these considerations are not taken into account in the clustering task. The main implicit argument is that these errors are zero mean and could be neglected at the first glance. The aim of this note is to design a new algorithm to perform clustering over contaminated datasets and to show that it can significantly improve the expected performances of a standard clustering algorithm which neglect this additional source of randomness.  
\paragraph{The $k$-means clustering problem.}
         The $k$-means is one of the most popular clustering method. The principle is to give an optimal partition of the data minimizing a distortion based on the Euclidean distance. The model of $k$-means clustering can be written as follows. Consider a random vector $X\in\R^d$ with law $P$ on a probability space $(\Omega,\mathcal{F},\P)$, and a number of clusters $k\geq 1$. Given $n$ i.i.d. copies $X_1,\ldots, X_n$ of $X$, we want to build a set of centers $\bc=(c_1,\ldots,c_k)\in\R^{dk}$ minimizing the distortion:
\beqn
\label{risk}
W_P(\bc):=\E_P\min_{j=1,\ldots, k}\no X-c_j\no^2,
\eeqn
where $\no\cdot\no$ stands for the Euclidean distance in $\R^d$. The existence of a  minimizer of (\ref{risk}) is guaranteed by \cite{graf}. In the rest of the paper, a minimizer of (\ref{risk}) is called an oracle and is denoted by $\bc^\star$. The problem of finite dimensional clustering becomes to estimate $\c^\star\in\R^{dk}$.

A natural way of minimizing (\ref{risk}) thanks to the collection $X_1, \ldots, X_n$ is to consider the empirical distortion:
\beqn
\label{empirisk}
W_{P_n}(\bc)=\frac{1}{n}\sum_{i=1}^n\min_{j=1,\ldots, k}\no X_i-c_j\no^2,
\eeqn
where $P_n:=(1/n)\sum\delta_{X_i}$ is the empirical measure. Thanks to an uniform law of large numbers, we can expected convergence properties for $\bc^\star_n:=\arg\min_{\bc}  W_{P_n}(\bc)$ to an oracle $\bc^\star$. In this direction, many authors have investigated the theoretical properties of $\bc^\star_n$. As a seminal example, \cite{pollard82} proves central limit theorem and consistency result under regularity conditions.

 In this direction, the basic iterative procedure of $k$-means was proposed by Lloyd in a seminal work (\cite{lloyd}, first published in 1957 in a Technical Note of Bell Laboratories). The algorithm is illustrated in Figure \ref{fig1}. The procedure calculates, from an initialization of $k$ centers, the associated Vorono\"i cells and actualize the centers with the means of the data on each Vorono\"i cell. 
\begin{figure}
\label{fig1}
---------------------------------------------------------------------------------------------------------------------
\begin{enumerate}
\item Initialize the centers $\mathbf{c}^{(0)}=(c_{1}^{(0)},\dots,c_{k}^{(0)})\in\mathbb{R}^{dk}$
\item Repeat until convergence:
\begin{enumerate}
\item Assign data points to closest clusters.
\item Re-adjust the center of clusters.
\end{enumerate}
\item Compute the final partition by assigning data points to the final
closest clusters $$\mathbf{\hat{c}_n}=(\hat{c}_{1},\dots,\hat{c}_{k}).$$
\end{enumerate}
---------------------------------------------------------------------------------------------------------------------

\caption{\textit{Lloyd algorithm.\label{fig:$k$-means algorithm.}}}

\end{figure}The $k$-means with Lloyd algorithm is considered as a staple in the study of clustering methods. The time complexity is approximately linear, and appears as a good algorithm for clustering spherical well-separated classes, such as a mixture of gaussian vectors. However, the principal limitation in the previous algorithm is that it only reaches a local minimum of the empirical distortion $W_{P_n}(\bc)$. Indeed, \cite{bubeck} has proved that $k$-means does Newton iterations in the sense that step 2.(b) in Figure \ref{fig1} corresponds exactly to a step of a Newton optimization. More precisely, the trajectories of two consecutives centers $\bc^t$ and $\bc^{t+1}$ visited by the algorithm satisfy:
$$
W_{P_n}(\bc^\alpha)\leq W_{P_n}(\bc^t),\,\forall \bc^\alpha=(1-\alpha)\bc^t+\alpha \bc^{t+1},\,\alpha\in (0,1).
$$
A natural consequence of this result is that if a local minimizer of the empirical distortion is reached, then the algorithm stops to this local optimum. 

\begin{figure}
\label{fig2}
\begin{center}
\includegraphics[width=14cm]{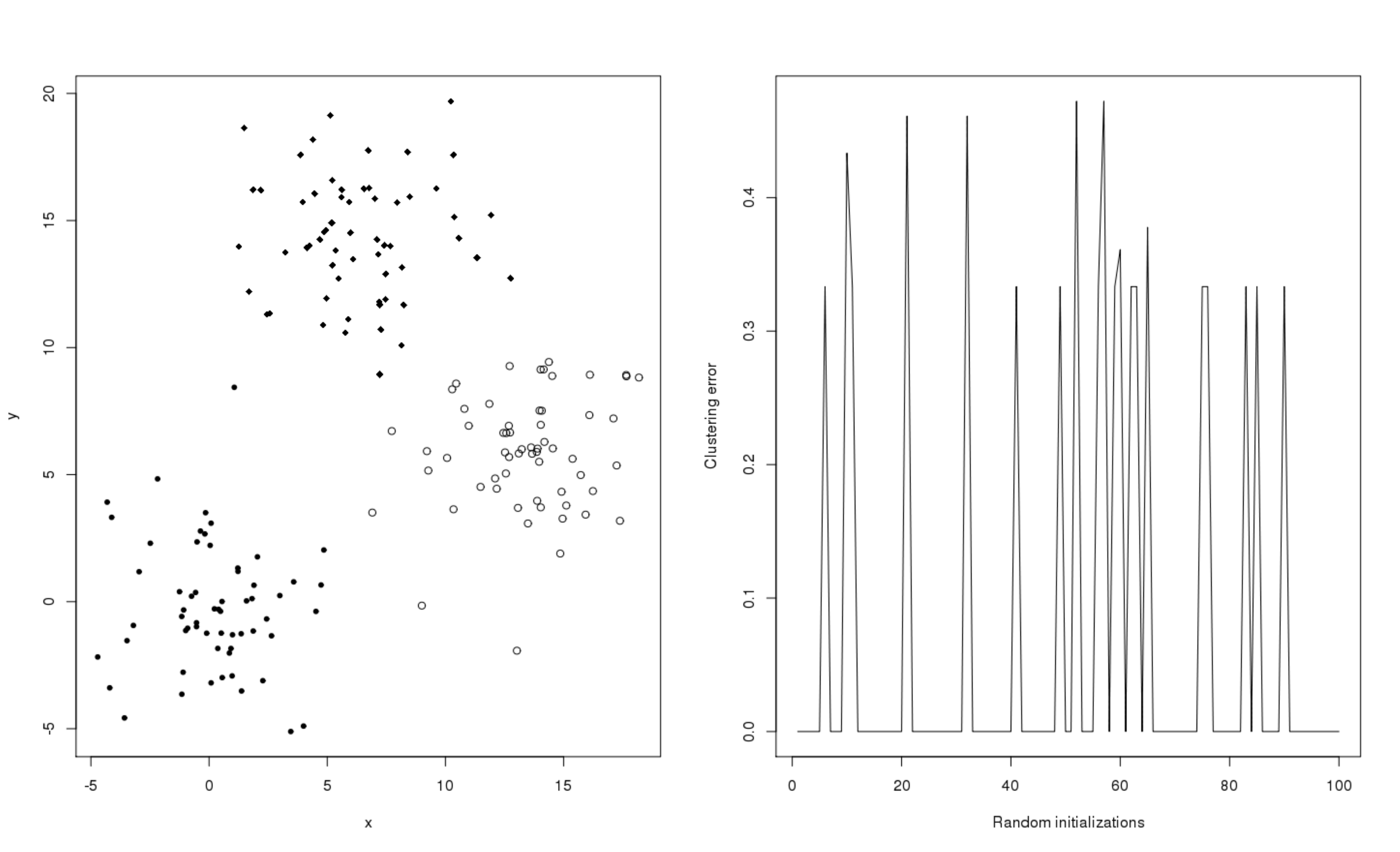}
\end{center}

\caption{Left panel shows the observations $X_1,\ldots, ,X_n$ from a mixture of three spherical gaussian (see Section \ref{exp} for the experimental set-up). The right panel shows the classification error of 100 runs of the Lloyd algorithm with random initialization. It shows how the initialization affects the performances of the $k$-means. \label{fig:exp2}}
\end{figure}

This principal drawback of the $k$-means algorithm of Lloyd is due to the non-convexity of the empirical distortion $\bc\mapsto W_{P_n}(\bc)$. This property appears practically on the dependence on the solution of the algorithm to the initialization. Different runs of the $k$-means with random initializations lead to unstable solutions. Figure \ref{fig:exp2} illustrates dramatically this phenomenon in a mixture of three spherical gaussians. 
\paragraph{The noisy $k$-means clustering problem.} In this paper, we are interested in the problem of noisy clustering (see \cite{aoinq}). The problem is still to minimize the distortion (\ref{risk}), but when we only have at our disposal a noisy sample:
$$
Z_i=X_i+\epsilon_i,\,i=1,\ldots ,n.
$$
Here, $\epsilon_i$, $i=1,\ldots ,n$ are i.i.d. random noise with density $\eta$ with respect to Lebesgue measure. As a result, the density of the observations $Z_1, \ldots, Z_n$ is the convolution product $f*\eta(x):=\int f(t)\eta(x-t)dt$. For this reason, we are facing an inverse statistical learning problem (see \cite{isl}). The empirical measure $P_n=1/n\sum\delta_{X_i}$ is not available and only a contaminated version $\tilde P_n:=1/n\sum\delta_{Z_i}$ is observable. 

As a result, the empirical distortion (\ref{empirisk}) is not computable. We can only study the empirical distortion with respect to the contaminated data $Z_1, \ldots, Z_n$, namely the quantity $W_{\tilde P_n}(\bc)$, for $\bc\in\R^{dk}$. Unfortunately, a standard minimization of $W_{\tilde P_n}(\bc)$ seems to fail since for any fixed codebook $\bc\in\R^{dk}$:
$$
\E W_{\tilde P_n}(\bc)=W_{P_Z}(\bc)=\int \min\no x-c_j\no^2 f*\eta(x) dx\not=W_P(\bc).
$$
This phenomenon can be interpreted as follows. If we use a basic clustering algorithm based on the minimization of the empirical distortion (such as the $k$-means algorithm) when we deal with noisy data, the expected criterion does not coincide with the distortion. This phenomenon gives rise to two different situations, which can be summarized as follows:
 \begin{figure}
\begin{center}
\includegraphics[width=14cm]{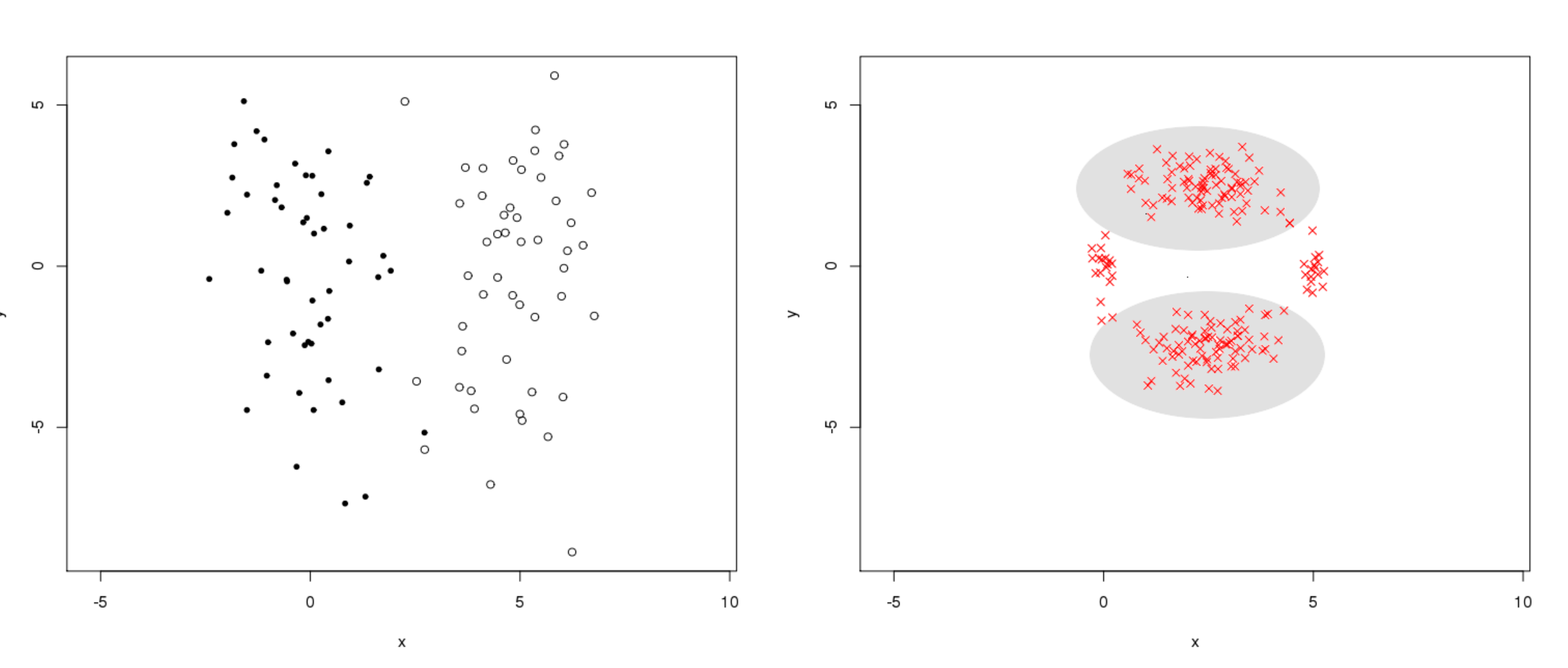}
\end{center}
\caption{The experimental set-up is detailed in Section \ref{s:exp1}.
Left panel shows an example of noisy dataset $Z_i=X_i+\epsilon_i, i=1, \ldots,n$ of two spherical gaussians (the $X_i$'s) with additive vertical noise (the $\epsilon_i$'s) (\textbf{Mod1($9$) in Section \ref{s:exp1}}.  Right panel shows solutions $\hat\bc_n$ of solutions over 100 runs of model \textbf{Mod1($9$)}. In most of the runs (grey ellipsoids in the right panel), the solutions  $\hat\bc_n$ propose an horizontal separation, corresponding to a bad clustering of direct inputs $X_i, i=1, \ldots n$ . \label{fig:exp1}}
\end{figure}
\begin{itemize}
\item At the first glance, the inequality $W_{P_Z}(\bc)\not=W_P(\bc)$  can be considered harmless if the following two oracle sets coincide:
\beqn
\label{sameoracle}
\arg\min_{\bc\in\R^{dk}} W_P(\bc)=\arg\min_{\bc\in\R^{dk}} W_{P_Z}(\bc).
\eeqn
Indeed, in this case, the global minimization of $\E W_{\tilde P_n}(\bc)$ lead coarselly to the best solution $\bc^\star$ thanks to an uniform law of large numbers. However, in practice, the global minimum is not available with standard Lloyd algorithm where only a local minimizer is guaranteed. As a result, even if the two oracle sets coincide in (\ref{sameoracle}), it will be more interesting to perform a noisy version of the well-known $k$-means (see Section \ref{s:exp2} for an illustration).
\item In the general case, there is no reason that (\ref{sameoracle}) holds. Indeed, if we consider an arbitrary mixture of random vectors for the distribution of $X$, a random additive noise can lead to different oracle $\bc^\star$ for the distortion (\ref{risk}). An illustration of such a framework is proposed in Section \ref{s:exp1}, where the $k$-means is not consistent (see also Figure \ref{fig:exp1}).
\end{itemize}
These considerations motivate the introduction of a clustering method which takes into account the law of the measurement errors. The noisy $k$-means algorithm will tackle this issue by using a deconvolution method.
\paragraph{Outlines.}
The paper is organized as follows. In Section \ref{model}, we present the theoretical foundations of the noisy $k$-means algorithm. This method is originated from the study of risk bounds in statistical learning with errors in variables. We present the construction of a noisy version of the standard $k$-means algorithm in Section \ref{noisyalgo}. This algorithm mixes a multivariate kernel deconvolution strategy based on Fast Fourier Transform (FFT) with the standard iterative Lloyd algorithm of the $k$-means. In Section \ref{exp}, we finally illustrate numerically the good efficiency of the noisy $k$-means algorithm to deal with noisy inputs. Section \ref{conclusion} concludes the paper with a discussion about the main challenging open problems.
\section{Theoretical foundations of noisy $k$-means}
\label{model}
The problem we have at hand is to minimize the distortion (\ref{risk}) thanks to an indirect set of observations $Z_i=X_i+\epsilon_i$, $i=1, \ldots, n$. This problem is a particular case of inverse statistical learning (see \cite{isl}) and is known to be an inverse (deconvolution) problem. As a result, we suggest to use a deconvolution estimator of the density $f$ in the standard $k$-means algorithm of Figure \ref{fig:$k$-means algorithm.}. For this purpose, consider a kernel $\mathcal{K}\in L_2(\R^d)$ such that $\mathcal{F}[\mathcal{K}]$ exists, where $\mathcal{F}$ denotes the standard Fourier transform with inverse $\mathcal{F}^{-1}$. Provided that $\mathcal{F}[\eta]$ exists and is strictly positive, a deconvolution kernel is defined as:
\begin{eqnarray}
\mathcal{K}_{\eta} & : & \R^d \to \R \nonumber \\
& & t \mapsto \mathcal{K}_\eta(t) = \FF^{-1}\left[ \frac{\FF[\mathcal{K}](\cdot)}{\FF[\eta](\cdot/\lambda)}\right](t).
\label{dk}
\end{eqnarray}
Given this deconvolution kernel, we introduce a deconvolution kernel estimator of the form:
\beqn
\label{decest}
\hat{f}_n(x)=\frac{1}{n}\sum_{i=1}^n\frac{1}{\lambda}\mathcal{K}_\eta\left(\frac{Z_i-x}{\lambda}\right),
\eeqn
where $\lambda\in\R^{d}_+$ is a bandwidth parameter. In the sequel, with a slight abuse of notations, we write $1/\lambda=(1/\lambda_1,\ldots, 1/\lambda_d)$. 

Originally presented in \cite{pinkfloyds}, the idea of Noisy $k$-means is to plug the deconvolution kernel estimator $\hat{f}_n(x)$ into the distortion (\ref{risk}). It gives rise to the following deconvolution empirical distortion:
\beqn
\label{noisyempirisk}
\tilde W_n(\bc)=\int_K \min_{j=1,\ldots, k}\no x-c_j\no^2 \hat{f}_n(x) dx,
\eeqn
where $\hat{f}_n(x)$ is the kernel deconvolution estimator (\ref{decest}) and $K\subset \R^{dk}$ is a compact domain. Finally, we denote by $\tilde\bc^\star_n$ the solution of the following stochastic minimization:
$$
\tilde\bc^\star_n=\arg\min_{\bc\in\R^{dk}}\tilde W_n(\bc).
$$
The statistical performances of $\tilde\bc^\star_n$ in terms of distortion (\ref{risk}) has been studied recently in \cite{aoinq}. It is based on uniform law of large numbers applied to the noisy empirical measure $\tilde P_n$. In particular, the consistency and the precise rates of convergence of the excess distortion can be stated as follows:  
\begin{theorem}[\cite{aoinq}]
\label{theory}
Given an integer $s\in\N^*$, suppose $f$ has partial derivatives up to order $s-1$, such that all the partial derivatives of order $s-1$ are lipschitz. Suppose Pollard's regularity assumption are satisfied (see \cite{pollard82}). Then, $\tilde\bc^\star_n$ with $\bar \lambda=n^{-1/(2s+2\sum_{v=1}^d\beta_v)}$ is consistent and satisfies:
$$
W_P(\tilde\bc^\star_{n})-W_P(\bc^\star)\leq C n^{-s/s+\sum_{v=1}^d\beta_v},
$$
where $\beta=(\beta_1,\ldots,\beta_v)$ is related with the asymptotic behaviour of the characteristic function of $\eta$ as follows:
$$
|\mathcal{F}[\eta](t)|\approx \Pi_{v=1}^d\left(\frac{t_v^2+1}{2}\right)^{- \beta_v/2}.
$$
\end{theorem}
\begin{remark}[Consistency]
Theorem \ref{theory} ensures the consistency of the stochastic minimization (\ref{noisyempirisk}) based on a noisy dataset $Z_i$, $i=1, \ldots, n$. The rates of convergence depend on the regularity of the density $f$ and $\eta$. The assumption over the smoothness of $f$ is a particular case of the more classical H\"older regularity. It is standard in the deconvolution literature (see for instance \cite{meister}) and more generally in the nonparametric statistical inference (see \cite{booktsybakov}). The assumption over the characteristic function of $\eta$ is also extensively used in the inverse problem literature (see for instance \cite{cavaliersurvey}).
\end{remark}
\begin{remark}[Bias-variance decomposition]
The proof of this result is based on a decomposition of the quantity $W_P(\tilde \bc^\star_n)-W_P(\bc^\star)$ into two terms: a bias term and a variance term. The variance term is controlled by using the theory of empirical processes, adapted to the noisy set-up (see \cite{isl}). The regularity assumption over the density $f$ allows us to control the bias term and to get the proposed rates of convergence.
\end{remark}
\begin{remark}[Choice of the bandwidth]
The result of Theorem \ref{theory} holds for a particular choice of $\hat f_n$ in (\ref{noisyempirisk}), namely with a particular bandwidth $\bar\lambda$ in (\ref{decest}) such that:
\beqn
\label{bandwidthchoice}
\bar \lambda=n^{-1/(2s+2\sum_{v=1}^d\beta_v)}.
\eeqn
This choice trades off the bias term and the variance term in the proof. It depends explicitely on the regularity of the density $f$, throught its H\"older exponent $s$. From practical viewpoint, a data-driven choice of the bandwidth $\lambda$ is an open problem (see \cite{loustauchichi} for a theoretical point of view).
\end{remark}
\section{Noisy $k$-means algorithm}
\label{noisyalgo}
When we consider direct
data $X_{1},\dots,X_{n}$, we want to minimize the empirical distortion associated with the empirical mesure $P_n$ defined in (\ref{empirisk}), over $\bc=(c_{1},\dots,c_{k})\in\R^{dk}$ the set of $k$ possible centers. This leads to the well-known $k$-means or Lloyd algorithm presented in Section \ref{sec:intro}. Similarly, in the noisy case, when considering indirect
data $Z_{1},\ldots,Z_{n}$, a deconvolution empirical distortion is 
defined as:
$$
\tilde W_n(\bc)=\frac{1}{n}\sum_{i=1}^{n}\int \min_{j=1,\dots,k}\left\Vert x-c_{j}\right\Vert ^{2}\hat{f}_n(x)dx.
$$
Reasonably, a noisy clustering algorithm could be adapted, following the direct case and the construction of the standard $k$-means. In this section, the purpose is two-fold: on the one hand, a clustering
algorithm for indirect data derived from first order conditions is
proposed. On the second hand, practical and computational considerations
of such an algorithm are discussed.
\subsection{First order conditions}
Let us consider an observed corrupted data sample $Z_{1},\dots,Z_{n}\in\mathbb{R}^{d}$
which is generated from the additive measurement error model of Section \ref{model} as follows: \begin{equation}
Z_{i}=X_{i}+\epsilon_{i},\,\,\forall i\in\left\{ 1,\dots,n\right\}.\label{eq:model}\end{equation}
The following theorem gives the first order conditions to minimize the empirical distortion (\ref{noisyempirisk}). In the sequel, $\nabla f(x)$ denotes the gradient of $f$ at point $x\in\R^{dk}$.
\begin{theorem}
\label{representer}
Suppose assumptions of Theorem \ref{theory} are satisfied. Then, for any $\lambda>0$:
\beqn
\label{eq:clusters_estimation}
\bc_{\ell, j}=\frac{\sum_{i=1}^{n}\int_{V_{j}}x_{\ell}\mathcal{K}_{\eta}\left(\frac{Z_{i}-x}{\lambda}\right)dx}{\sum_{i=1}^{n}\int_{V_{j}}\mathcal{K}_{\eta}\left(\frac{Z_{i}-x}{\lambda}\right)dx},\,\forall \ell\in\{1,\dots,d\}\,,\forall j\in\{1,\dots,k\}\Rightarrow \nabla \tilde W_n(\bc)=0_{\R^{dk}},
\eeqn
where $\bc_{\ell,j}$ stands for the $\ell$-th coordinates of the $j$-th centers, whereas $V_{j}$ is the Vorono\"i cell associated with center $j$ of $\bc$:
$$
V_{j}=\{x\in\R^d:\min_{u=1,\ldots,k}\no x-c_{u}\no=\no x-c_{j}\no\}.
$$
\end{theorem}
\begin{remark}[Comparison with the $k$-means]
It is easy to see that a similar result is available in the direct case of the $k$-means. Indeed, a necessary condition to minimize the standard empirical distortion $W_n(\cdot)$ is as follows:
$$
\bc_{\ell, j}=\frac{\sum_{i=1}^{n}\int_{V_{j}}x_{\ell}\delta_{X_i}dx}{\sum_{i=1}^{n}\int_{V_{j}}\delta_{X_i}dx},\,\forall \ell\in\{1,\dots,d\}\,,\forall j\in\{1,\dots,k\},
$$
where $\delta_{X_i}$ is the Dirac function at point $X_i$. Theorem \ref{representer} proposes a same kind of condition in the errors-in-variable case replacing the Dirac function by a deconvolution kernel.
\end{remark}
\begin{remark}[A simpler representation]
We can remark that by switching the integral with the sum in equation~(\ref{eq:clusters_estimation}),
the first order conditions on $\bc$ can be rewritten as follows :
\begin{equation}
\label{eq:estimClass}
\bc_{\ell,j}=\frac{\int_{V_{j}}x_{\ell}\hat{f}_n(x)dx}{\int_{V_{j}}\hat{f}_n(x)dx},\,\forall \ell\in\{1,\dots,d\},\,\forall j\in\{1,\dots,k\},
\end{equation}
where $\hat{f}_n(x)=1/n\sum_{i=1}^{n}\frac{1}{\lambda}\mathcal{K}_{\eta}\left(\frac{Z_{i}-x}{\lambda}\right)$
is the kernel deconvolution estimator of the density $f$. This property is at the core of the algorithm presented in Section \ref{sec:algo}.
\end{remark}
\begin{proof}
The proof is based on the first order conditions for the deconvolution empirical distortion defined in (\ref{noisyempirisk}) as:
$$
\tilde W_{n}(\mathbf{c})=\frac{1}{n}\sum_{i=1}^{n}{\displaystyle \int_{K}}\min_{j=1,\dots,k}\left\Vert x-c_{j}\right\Vert ^{2}\frac{1}{\lambda}\mathcal{K}_{\eta}\left(\frac{Z_{i}-x}{\lambda}\right)dx.
$$
Let us introduce the quantity $J(\mathbf{c},z)$ defined as:\[
J(\mathbf{c},z)={\displaystyle \int_{K}}\min_{j=1,\dots,k}\left\Vert x-c_{j}\right\Vert ^{2}\frac{1}{\lambda}\mathcal{K}_{\eta}\left(\frac{z-x}{\lambda}\right)dx.\]
For a fixed $z\in\R$, and for any $\bc,\bc'\in\mathbb{R}^{dk}$, let us consider the directional derivative
of the function $J(\cdot,z):\,\mathbb{R}^{dk}\rightarrow\mathbb{R}$,
at $\bc$ along the direction $\bc'$ defined as:
$$
\nabla_{\bc'}J(\bc,z)=\lim_{h\rightarrow0}\frac{J(\bc+\bc'h,z)-J(\bc,z)}{h}.
$$
Using simple algebra, we have, denoting $V_j$ the Vorono\"i cell associated to $c_j$ and $V_j(h)$ the Vorono\"i cell associated with $(\bc+h\bc')_j$:
\beqnn
&&J(\bc+\bc'h,z)-J(\bc,z)  =   \int_{K}\left[\min_{j=1,\dots,k}\left\Vert x-(\bc+\bc'h)_{j}\right\Vert ^{2}-\min_{j=1,\dots,k}\left\Vert x-c_{j}\right\Vert ^{2}\right]\frac{1}{\lambda}\mathcal{K}_{\eta}\left(\frac{z-x}{\lambda}\right)dx\\
&=&\sum_{j=1}^k\left[\int_{V_j\cap V_j(h)}\left(h^2\Vert c'_{j}\Vert^2-2h\langle x-c_j,c'_j\rangle\right)\frac{1}{\lambda}\mathcal{K}_\eta\left(\frac{z-x}{\lambda}\right)dx\right]+\int_{V(h)^C}r(\bc,\bc',x,h,\lambda)dx,
\eeqnn
where:
$$
V(h)=\bigcup_{j=1}^k \left(V_j\cap V_j(h)\right),
$$
and $x\mapsto r(\bc,\bc',x,h,\lambda)$ is a bounded function whose precise expression is not useful. Indeed, using dominated convergence and the fact that for any $x\in K$, there exists some $h(x)>0$ such that for any $h\leq h(x)$, $\ind_{V(h)^C}(x)=0$, we arrive at:
$$
\nabla_{\bc'}J(\bc,z)=\sum_{j=1}^k\int_{V_j}-2\langle x-c_j,c'_j\rangle\frac{1}{\lambda}\mathcal{K}_\eta\left(\frac{z-x}{\lambda}\right)dx.
$$
For $\bc'\in\left\{ e_{ij}=(0,\ldots, 0, 1,\ldots, 0)|i=1\dots d,j=1\dots k\right\} $ the canonical
basis of $\R^{dk}$, one has:
\[
\nabla_{e_{ij}}J(\bc,z)=-2\int_{V_{j}}( x_{i}-c_{ij}) \frac{1}{\lambda}\mathcal{K}_{\eta}\left(\frac{z-x}{\lambda}\right)dx.
\]
Then a sufficient condition on $\bc$ to have  $\nabla_{e_{\ell,j}}\sum_{i=1}^nJ(\bc,Z_i)=0$ is:
\begin{eqnarray}
c_{\ell ,j}=\frac{1/n\sum_{i=1}^{n}\int_{V_{j}}x_{\ell}\frac{1}{\lambda}\mathcal{K}_{\eta}\left(\frac{Z_{i}-x}{\lambda}\right)dx}{1/n\sum_{i=1}^{n}\int_{V_{j}}\frac{1}{\lambda}\mathcal{K}_{\eta}\left(\frac{Z_{i}-x}{\lambda}\right)dx}.\label{eq:}\end{eqnarray}
\end{proof}

\subsection{The noisy $k$-means algorithm}
\label{sec:algo}
In the same spirit of the $k$-means algorithm of Figure \ref{fig:$k$-means algorithm.}, we derive therefore an iterative algorithm, named Noisy $k$-means, which enables to find
a reasonable partition of the direct data from a corrupted sample. The noisy $k$-means algorithm consists in two steps : (1) a deconvolution estimation step in order to estimate the density $f$ of direct data from the corrupted data and (2) an iterative Newton's procedure according to~(\ref{eq:estimClass}). This second step can be repeated several times until a stable solution is available. 
\subsubsection{Estimation step}
In this step, the purpose is to estimate the density $f$ from the
model~(\ref{eq:model}) in which the $X_{1},\dots,X_{n}$ are unobserved.
Let us denote by $f_{Z}$ the density of corrupted data $Z$. Then,
according to~(\ref{eq:model}), $f_{Z}$ is the convolution product
of the densities $f$ and $\eta$ denoted by $f_{Z}=f*\eta$. Consequently, the following
relation holds : $\mathcal{F}[f]=\mathcal{F}[f_{Z}]/\mathcal{F}[\eta]$.
A natural property for the Fourier transform of an estimator $\hat f$ can be deduced:
\beqn
\label{eq:FT_densX}
\mathcal{F}[\hat{f}] =\widehat{\mathcal{F}}[f_{Z}]/\mathcal{F}[\eta],
\eeqn
where $\widehat{\mathcal{F}}[f_{Z}](t)=1/n\sum_{i=1}^{n}e^{i\left\langle t,Z_{i}\right\rangle }$
is the Fourier transform of the data. These considerations explain the introduction of the deconvolution kernel estimator (\ref{decest}) presented in Section \ref{theory}.
\begin{figure}
---------------------------------------------------------------------------------------------------------------------
\begin{enumerate}
\item Initialize the centers $\mathbf{c}^{(0)}=(c_{1}^{(0)},\dots,c_{k}^{(0)})\in\mathbb{R}^{dk}$
\item Estimation step:
\begin{enumerate}
\item Compute the deconvoluting Kernel $\mathcal{K}_{\eta}$ and its FFT
$\mathcal{F}(\mathcal{K}_{\eta})$.
\item Build a histogram of 2-d grid using linear binning rule and compute
its FFT: $\mathcal{F}(\hat{f}_{Z})$.
\item Compute: $\mathcal{F}(\hat{f})=\mathcal{F}(\mathcal{K}_{\eta})\mathcal{F}(\hat{f}_{Z})$.
\item Compute the Inverse FFT of $\mathcal{F}(\hat{f})$ to obtain the density
estimated of X: $\hat{f}=\mathcal{F}^{-1}(\mathcal{F}(\hat{f}))$.
\end{enumerate}
\item Repeat until convergence:
\begin{enumerate}
\item Assign data points to closest clusters in order to compute the Voronoi
diagram.
\item Re-adjust the center of clusters with equation~(\ref{eq:estimClass}).
\end{enumerate}
\item Compute the final partition by assigning data points to the final
closest clusters $\mathbf{\hat{c}}=(\hat{c}_{1},\dots,\hat{c}_{k})$.
\end{enumerate}
---------------------------------------------------------------------------------------------------------------------
\caption{The algorithm of Noisy $k$-means.\label{fig:Noisykmeans-algorithm.}}
\end{figure}In practice, deconvolution estimation involves $n$ numerical integrations
for each grid where the density needs to be estimated. Consequently,
 a direct programming of such a problem
is time consuming when the dimension $d$ of the problem increases. In order to speed the procedure, we have used the
Fast Fourier Transform (FFT). In particular, we have adapted the FFT
algorithm for the computation of multivariate kernel estimators proposed
by~\citep{Wand94} to the deconvolution problem. Therefore, the FFT
of the deconvoluting kernel is first computed.
Then, the Fourier transform of data $\widehat{\mathcal{F}}[f_{Z}]$ is
computed via a discrete approximation: an histogram on a grid
of $2$ dimensional cells is built before applying the FFT as it was
proposed in~\citep{Wand94}. Then, the discrete Fourier transform
of $f$ is obtained from equation~(\ref{eq:FT_densX}) and an estimation
of $f$ is found by an inverse Fourier transform.

\subsubsection{Newton's iterations step}

The center of the $j$th group on the $\ell$th component can therefore
be computed from~(\ref{eq:estimClass}) as follows~:\[
c_{ij}=\frac{\int_{V_{j}}x_{\ell}\hat{f}_n(x)dx}{\int_{V_{j}}\hat{f}_n(x)dx},\]
where $V_{j}$ stands for the Voronoi cell of the group $j$. \\

Consequently, the estimation procedure needs two different steps : an estimation step to compute the kernel density estimator, and an iterative procedure to converge to the first order conditions. The noisykmeans algorithm
is summed up in Figure~\ref{fig:Noisykmeans-algorithm.}.
\section{Experimental validation}
\label{exp}
Evaluation of clustering algorithms is not an easy task (see \cite{scienceorart}). In this section, we choose to highlight some important phenomena related with the inverse problem we have at hand. These phenomena are of different nature and show the usefulness of the deconvolution step when we deal with noisy data:
\begin{itemize}
\item In the first experiment, we show that the noisy $k$-means algorithm is consistent to discriminate two spherical well-separated gaussian in two dimension, when we observe corrupted sample with increasing variance. It illustrates the particular case of Section \ref{sec:intro} (Figure \ref{fig:exp1}) where Noisy $k$-means appears as a good alternative to the standard $k$-means.
\item  The second experiment is related with Figure \ref{fig:exp2} of Section \ref{sec:intro}, where the initialization affects the $k$-means. In this case, by decreasing the distance between the 3 spherical gaussians, the Noisy $k$-means algorithm highlights a good resistency.
\end{itemize}
\subsection{First experiment}
\label{s:exp1}
\subsubsection{Simulation setup} In this simulation, we consider, for $u\in\{1,\ldots,  10\}$, the following model, called \textbf{Mod1($u$)}:
$$
\label{modexp1}
Z_i=X_i+\epsilon_i(u),\,i=1, \ldots, n,\,\,\,\,\textbf{Mod1($u$)}
$$
where:
\begin{itemize}
\item $(X_i)_{i=1}^n$ are i.i.d. with density $f=1/2f_{\mathcal{N}\left(0_2,I_2\right)}+1/2f_{\mathcal{N}\left((5,0)^T,I_2\right)}$ 
\item and $(\epsilon_i(u))_{i=1}^n$ are i.i.d. with law $\mathcal{N}\left(0_2,\Sigma(u)\right)$, where $\Sigma(u)$ is a diagonal matrix with diagonal vector $(0,u)^T$, for $u\in\{1,\ldots, 10\}$. 
\end{itemize}
In this case, the error is concentrated into the vertical axe, and increases with parameter $u\in\{1,\ldots, 10\}$.

\begin{figure}
\begin{center}
\includegraphics[width=8cm]{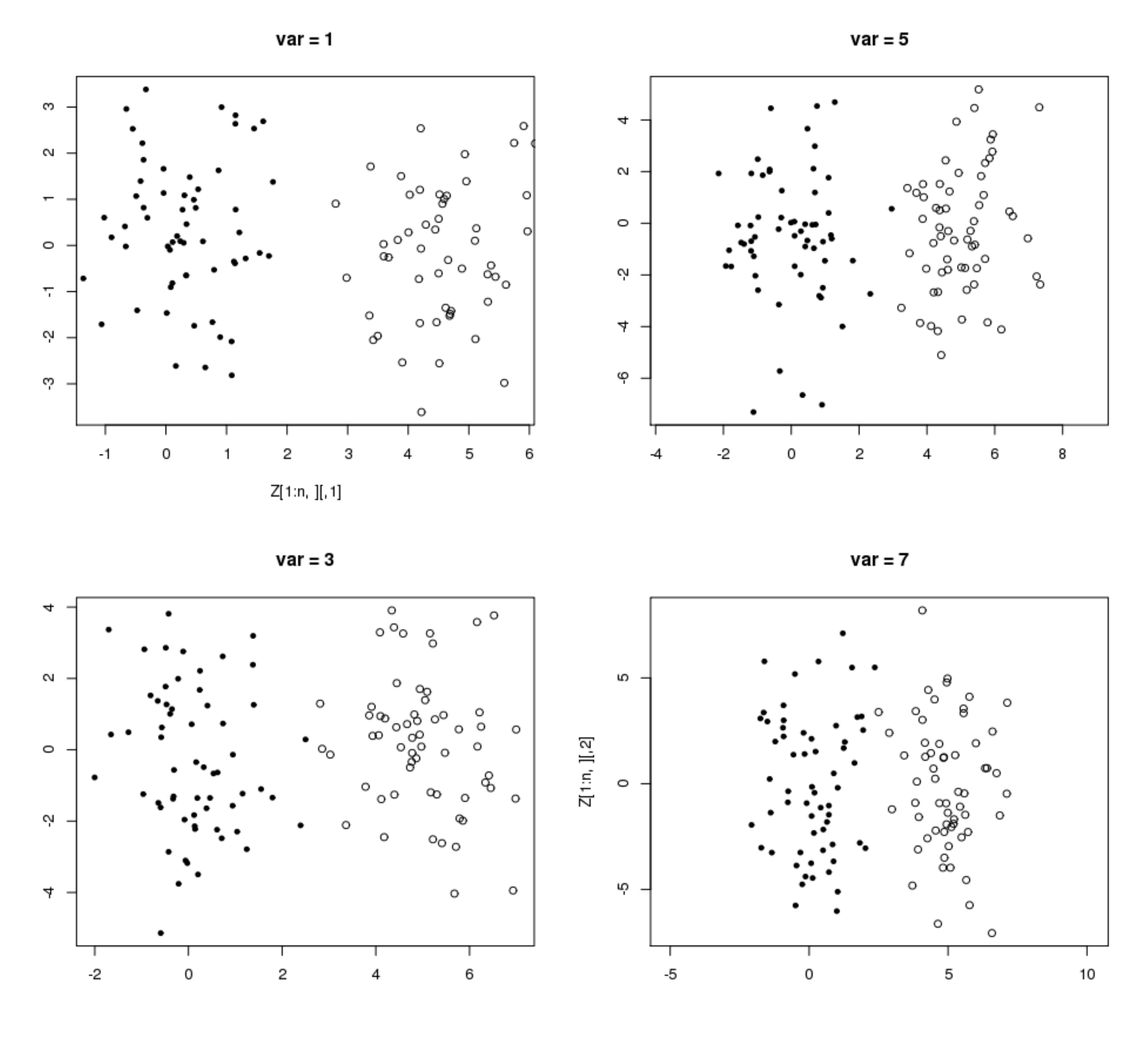}
\end{center}

\caption{First experiment's setting for $u\in\{1,3,5,7\}$ \label{exampleexp1}}

\end{figure}

We study the behaviour of the Lloyd algorithm of Figure \ref{fig1} and the noisy $k$-means algorithm of Figure~\ref{fig:Noisykmeans-algorithm.} in \textbf{Mod1($u$)}, for $u\in\{1,\ldots, 10\}$. For this purpose, for each $u$, we run 100 realizations of training set $\{Z_1,\ldots, Z_n\}$ from \textbf{Mod1($u$)} with $n=100$. At each realization, we run Lloyd algorithm and Noisy $k$-means with the same random initialization. The value of $\lambda>0$ in Noisy $k$-means is tuned with a grid $\Lambda$ of $20\times 20$ parameters as follows:
$$
\hat\lambda=\arg\min_{\lambda\in\Lambda}\sum_{t=1}^{1000}\min_{j=1,2}\no X^{\mathrm{tun}}_t-(\tilde \bc_n)_j\no^2,
$$ 
where $\tilde \bc_n=((\tilde \bc_n)_1,(\tilde \bc_n)_2)$ is the solution of Noisy $k$-means with parameter $\lambda=(\lambda_1,\lambda_2)$ and $(X^{\mathrm{tun}}_t)_{t=1}^{1000}$ is an additional i.i.d. sample with density $f$.\\ In Figure~\ref{figexp1}, we are mainly interested in the clustering risk at each realization, defined as:
\beqn
\label{error}
r_n(\hat \bc)=\frac{1}{100}\sum_{i=1}^{100}\ind_{Y_i\not=\hat  \bc(X_i)},
\eeqn
where $\hat\bc$ denotes either the standard Lloyd algorithm performed on the dataset $\{Z_1,\ldots ,Z_n\}$  or the Noisy $k$-means of Figure~\ref{fig:Noisykmeans-algorithm.} with $\lambda=\hat\lambda$.
\subsubsection{Results of the first experiment}
\begin{figure}
\begin{center}
\includegraphics[width=20cm,angle=90]{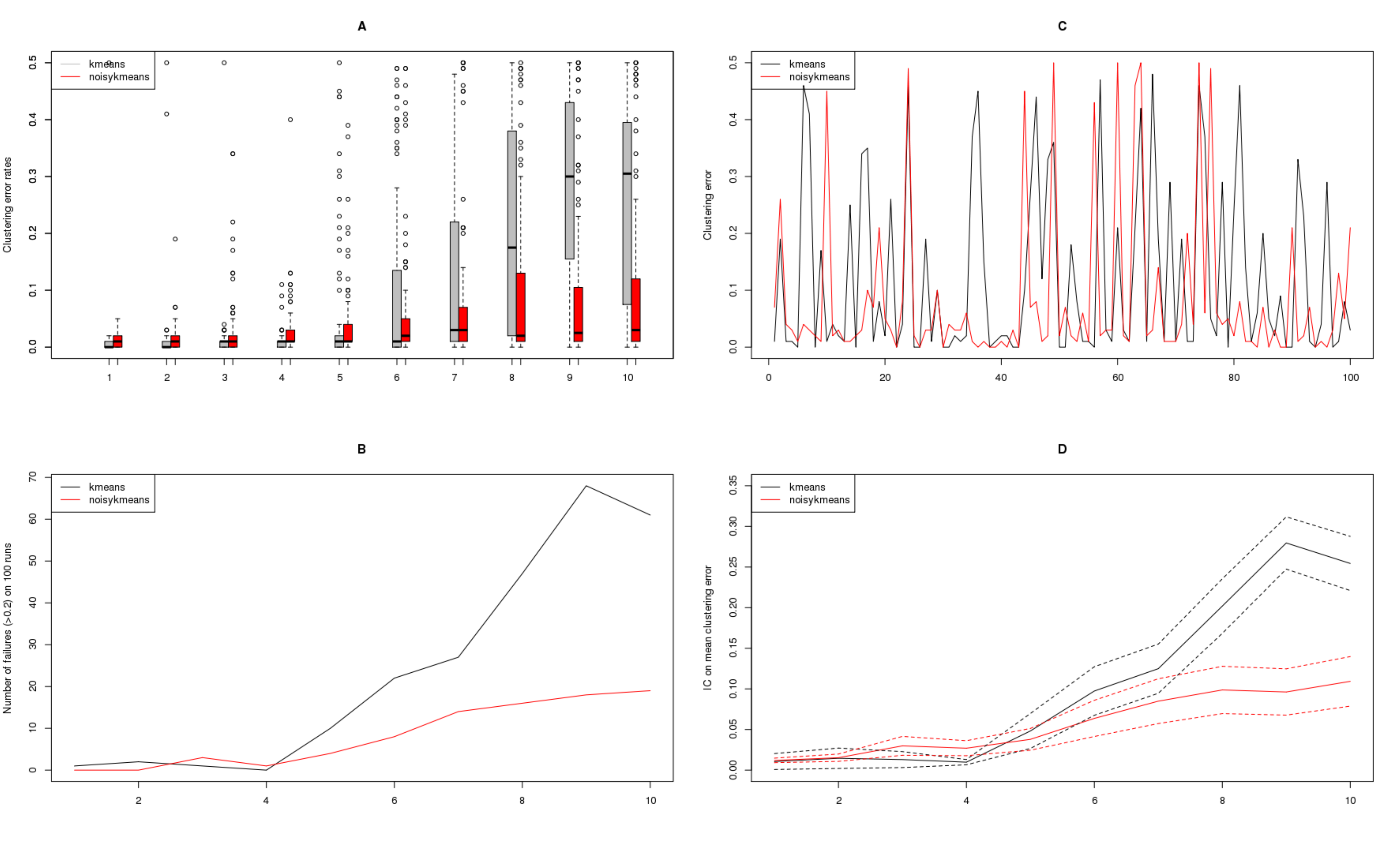}
\end{center}

\caption{Results of the first experiment \label{figexp1}}

\end{figure}
Figure \ref{figexp1} illustrates the result of the first experiment. At first, it shows rather well the lack of efficiency of the standard $k$-means when we deal with errors in variables. When the variance of the noise $\epsilon$ increases, the performances of the $k$-means are deteriorated. On the contrary, the noisy $k$-means shows a good robustness to this additional source of noise. Here is a detailed explanation of Figure \ref{figexp1}.
\paragraph{A} These boxplots show the evolution of the clustering risk (\ref{error}) of the two algorithms when the variance increases. When parameter $u\in\{1,\ldots, 5\}$, the results are comparable and standard $k$-means seems to slightly outperform Noisy $k$-means. However, when the level of noise in the vertical axe becomes higher (i.e. $u\geq 6$), Lloyd algorithm shows a very bad behaviour. On the contrary, noisy $k$-means seems to be more robust in these situations.
\paragraph{B} Here, we are interested on situations where the studied algrithms fail, i.e. when the clustering errors $r_n(\hat \bc)>0.2$. Figure 5.B shows rather well the robustness of the noisy $k$-means in comparison with the standard $k$-means. The situation becomes problematic for $k$-means when $u\geq 6$ where the numbers of failures is bigger than $20$ (from 22 to 68 over a total of 100 runs). On the contrary, the maximum of failures of the Noisy $k$-means does not exceed $19$. 
\paragraph{C} Figure 6.C is a precise illustration of the behaviour of the two algorithms in the particular model \textbf{Mod1($7$)}. We have plot the clustering errors $r_n(\hat \bc)$ for each run in this model. Of course, here again, Noisy $k$-means outperforms standard $k$-means at many runs. However, at some runs, the standard $k$-means does a good job whereas Noisy $k$-means completely fails. This can be explained by the dependence on the solution to the random initialization (and then on the non-convexity of the problem).
\paragraph{D} Finally, last plot deals with the mean clustering risk in each model \textbf{Mod1($u$)}, $u\in\{1,\ldots, 10\}$ and the corresponding confidence intervals, calculated thanks to the following formula:
$$
\left[\mu(r_n(\hat \bc))-1.96\times \frac{\sigma(r_n(\hat \bc))}{\sqrt{n}},\mu(r_n(\hat \bc))-1.96\times \frac{\sigma(r_n(\hat \bc)}{\sqrt{n}}\right],
$$
where $\mu(r_n(\hat \bc))$ (and respectively $\sigma(r_n(\hat \bc))$) is the mean clustering risk over the 100 runs (and respectively the standard deviation).

This study highlights the robustness of the Noisy $k$-means in comparison with the $k$-means. Indeed, the associated IC are well-separated when $u\geq 7$.
\subsection{Second experiment}
\label{s:exp2}
\subsubsection{Simulation setup} 
\begin{figure}
\begin{center}
\includegraphics[width=8cm]{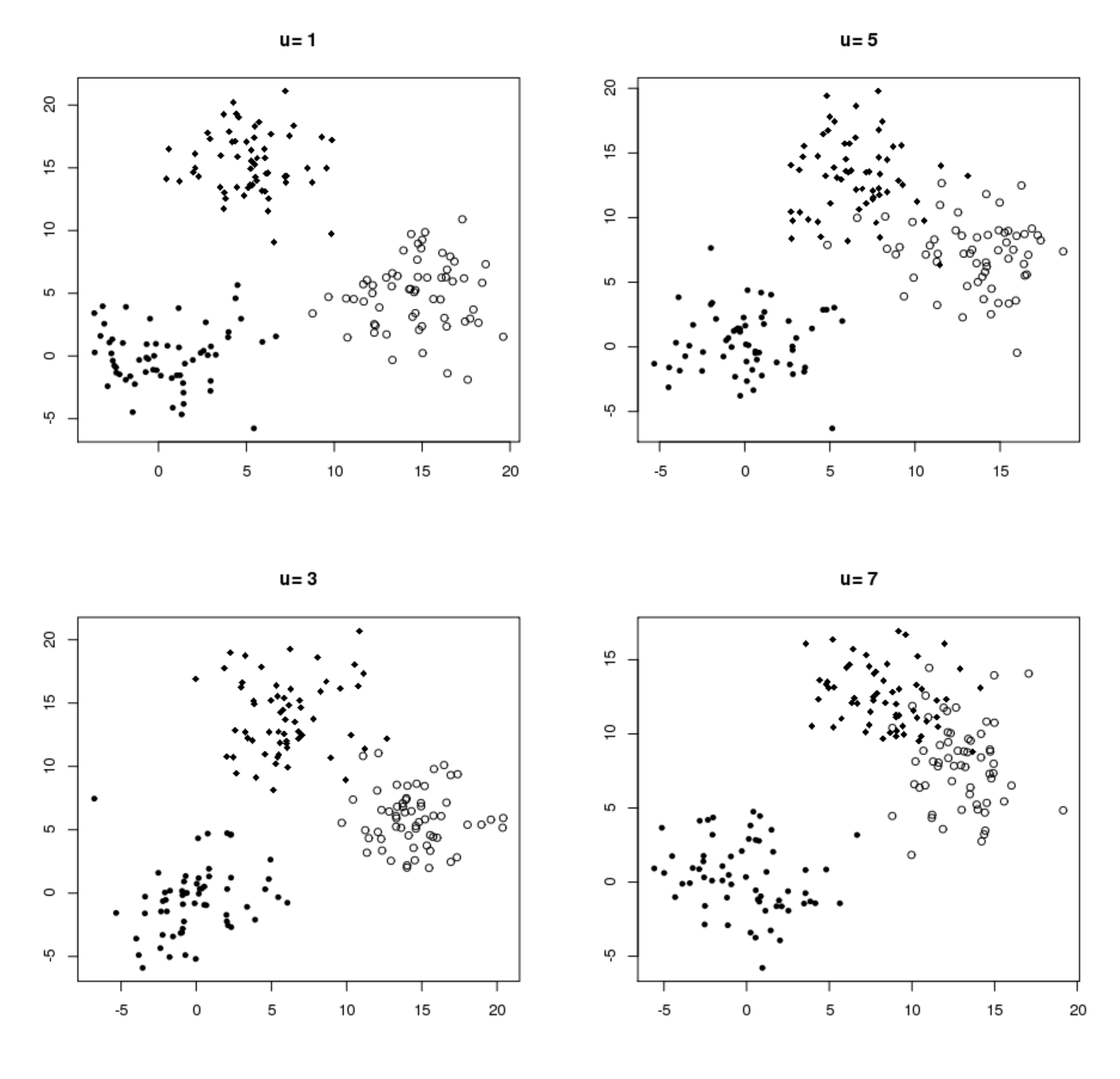}
\end{center}

\caption{Second experiment's setting for $u\in\{1,3,5,7\}$ \label{exampleexp2}}

\end{figure}

In this simulation, we consider, for $u\in\{1,\ldots,  10\}$, the model \textbf{Mod2($u$)} as follows:
$$
\label{modexp2}
Z_i=X_i(u)+\epsilon_i,\,i=1, \ldots, n,\,\,\,\,\textbf{Mod2($u$)}
$$
where:
\begin{itemize}
\item $(X_i(u))_{i=1}^n$ are i.i.d. with density $$
f=1/3f_{\mathcal{N}\left(0_2,I_2\right)}+1/3f_{\mathcal{N}\left((a,b)^T,I_2\right)}+1/3f_{\mathcal{N}\left((b,a)^T,I_2\right)},
$$
where $(a,b)=(15-(u-1)/2,5+(u-1)/2)$, for $u\in\{1,\ldots, 10\}$,
\item and $(\epsilon_i)_{i=1}^n$ are i.i.d. with law $\mathcal{N}\left(0_2,\Sigma\right)$, where $\Sigma$ is a diagonal matrix with diagonal vector $(5,5)^T$.
\end{itemize}
In this case, the errors in variables is stable but the distance between the clusters is decreasing (see Figure \ref{exampleexp2}).

As in the first experiment, we study the behaviour of both algorithms in \textbf{Mod2($u$)}, for $u\in\{1,\ldots, 10\}$. For this purpose, for each $u$, we run 100 realizations of training set $\{Z_1,\ldots, Z_n\}$ from \textbf{Mod2($u$)} with $n=180$. At each realization, we run Lloyd algorithm and Noisy $k$-means with the same random initialization and the same tuned choice of $\hat \lambda>0$ described in Section \ref{s:exp1}.
\subsubsection{Results of the second experiment}
\begin{figure}
\begin{center}
\includegraphics[width=20cm,angle=90]{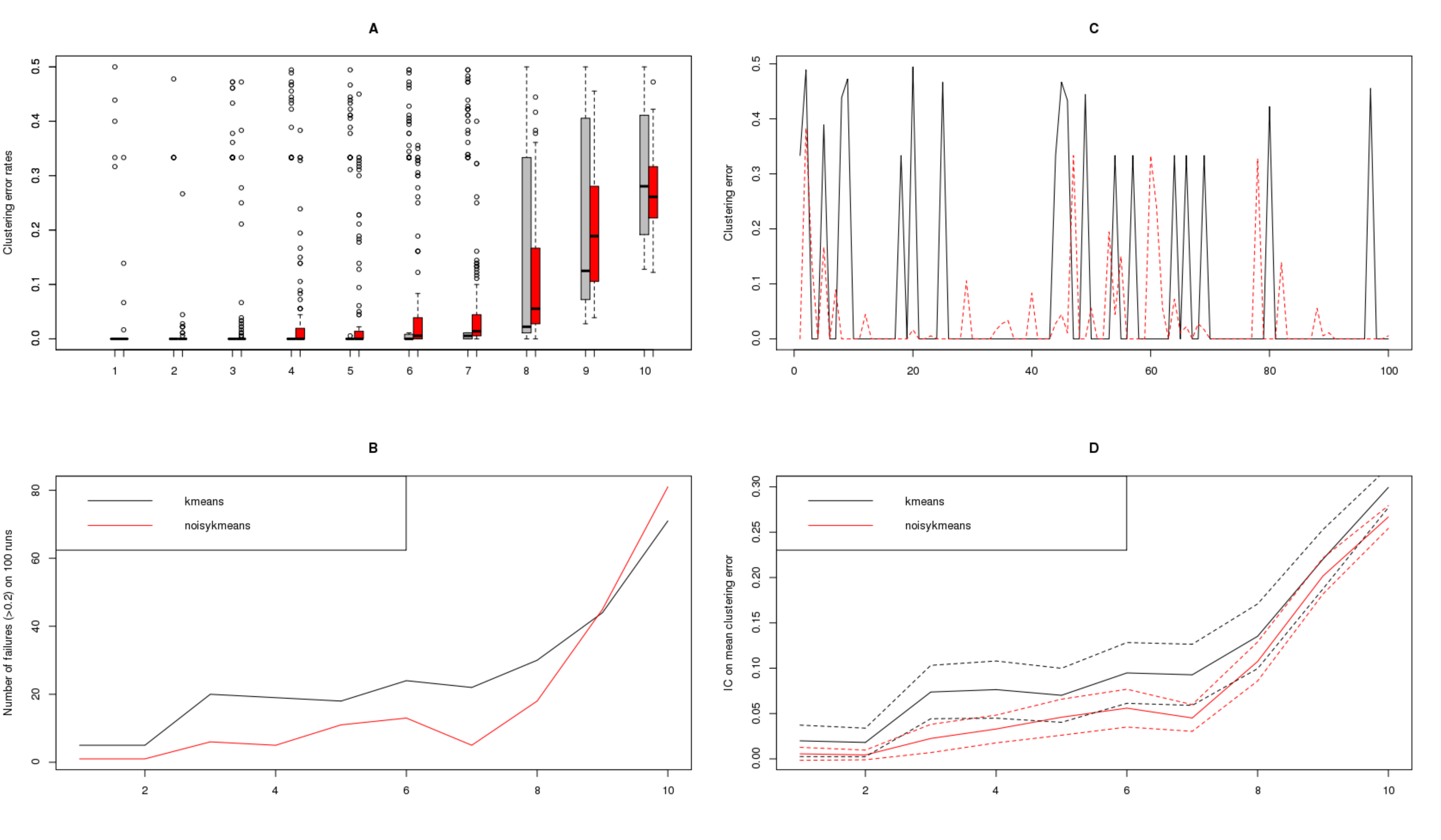}
\end{center}

\caption{Results of the second experiment \label{figexp2}}

\end{figure}

Figure \ref{figexp2} illustrates the result of the second experiment. At first, it shows rather well the difficulty of the problem when the distance between the clusters decreases. When parameter $u$ increases, the performances are deteriorated. However, the noisy $k$-means shows a better robustness for this problem. Here is a detailed explanation of Figure \ref{figexp2}.
\paragraph{A} These boxplots show the evolution of the clustering risk (\ref{error}) of the two algorithms when parameter $u$ increases. When $u\in\{1,\ldots, 7\}$, the results are comparable and standard $k$-means seems to outperform slightly Noisy $k$-means. However, when $u\geq 8$, Lloyd algorithm shows a very bad behaviour. On the contrary, Noisy $k$-means seems to be more robust in these difficult situations.
\paragraph{B} Here, we are interested on situations where the studied algrithms fail, i.e. when the clustering errors $r_n(\hat \bc)>0.2$. Figure 5.B shows rather well the better performances of the Noisy $k$-means in comparison with the standard $k$-means when $u\leq 8$. However, the number of failures becomes problematic for $k$-means and Noisy $k$-means when $u\geq 9$ where the numbers of failures is bigger than $30$ (over a total of 100 runs). It corresponds to very difficult clustering problems.
\paragraph{C} Figure 8.C is a precise illustration of the behaviour of the two algorithms in the particular model \textbf{Mod2($4$)}. We have plot the clustering errors $r_n(\hat \bc)$ for each run in this model. Here, Noisy $k$-means outperforms standard $k$-means at many runs. However, at some runs, the standard $k$-means does a good job whereas Noisy $k$-means seems to fail. This can be explained by the dependence on the solution to the random iteration (and then on the non-convexity of the problem).
\paragraph{D} Finally, last plot deals with the mean clustering risk in each model \textbf{Mod2($u$)}, $u\in\{1,\ldots, 10\}$ with the corresponding confidence intervals (see the previous subsection for the precise formula). With such statistics, the ability of noisy $k$-means seems to be clearly better than $k$-means, for any value of $u\in\{1,\ldots, 10\}$.
\subsection{Conclusion of the experimental study}
This experimental study can be seen as a first step into the study of clustering from a corrupted data. The results of this section are quiet promising and show rather well the importance of the deconvolution step in this inverse statistical learning problem. The first experiments show that standard $k$-means is not abble to separate two spherical gaussians in the presence of an additive vertical noise. In this case, noisy $k$-means appears as an interesting solution, mainly when this noise is increasing. Moreover, we also show that in the presence of three spherical gaussians with different separations, noisy $k$-means outperforms $k$-means.
\section{Conclusion}
\label{conclusion}
This technical report presents a new algorithm to deal with clustering with errors in variables. The procedure mixes deconvolution tools with standard $k$-means iterations. The design of the algorithm is based on the calculus of the first order conditions of a deconvolution empirical distortion, based on a deconvolution kernel. As a result, the algorithm can be seen as the indirect counterpart of the Lloyd algorithm of the direct case, which appears to do exactly Newton's iterations (see \cite{bubeck}). Due to the deconvolution step, the algorithm extensively uses the two-dimensional FFT (Fast Fourier Transform) algorithm.

We show numerical results in particular simulated examples to illustrate various phenomena. At first, we show that the standard $k$-means is not abble to separate two spherical gaussian in the presence of a vertical noise. On the contrary, noisy $k$-means can deal with measurement errors thanks to the deconvolution step. Moreover, we show that noisy $k$-means is also more suitable to separate three spherical gaussians with different separations.\\

This algorithm could be considered as a staple into the study of noisy clustering, or more generally classification with errors in variables. As the popular $k$-means, it suffers from non-convexity and as the popular $k$-means, the initialization affects the performances. Moreover, due to the inverse problem we have at hand, the dependence on the bandwidth $\lambda>0$ has to be considered seriously. In this paper, we perform the algorithm with a tuning choice of the bandwidth, where the criterion to choose the bandwidth depends on the density $f$ itself. Of course in practice this choice is not available and the problem of choosing the bandwidth is a challenging future direction. However, this problem is not out of reach. Indeed, recently, \cite{loustauchichi} proposes an adaptive choice of the bandwidth based on the Lepski's procedure.  

Another problem of the algorithm of this paper is the dependence on the law of the noise $\epsilon$. The construction of the deconvolution kernel needs the entire knowledge of the density $\eta$, which is used in the algorithm of noisy $k$-means. This problem is very popular in the statistical inverse problem literature, where various solutions are proposed. The most classical one is to use repeated measurements to estimate the law of $\epsilon$. In this direction, we have compiled an adaptive (to the noise) algorithm to deal with repeated measurements. This algorithm estimates the Fourier transform of $\eta$ thanks to the repeated measurements. We omit this part for concisions.\\

Finally, applications to real-datasets is still in progress. To the best of our knowledge, the problem of clustering with noisy data is rather new and benchmark datasets are not easily available. However, there is nice hope that existing and popular datasets could be considered in future works. In this paper, we highlight good robustness for noisy $k$-means when the different clusters are not well separated. An interesting direction for future works is to consider difficult datasets which can be fitted to the model of clustering with errors in variables. Indeed ,to perform the algorithm, the question is the following : is it a model of clustering with errors in variables ? Of course, the answer is not possible without a detailed knowledge of the experimental set-up and eventually repeated measurements. We argue that this knowledge could be a way of improving classification rates for many problems.

         \bibliographystyle{plain}
         \bibliography{referencejmlr}
\end{document}